\documentclass{article}

\usepackage[nonatbib, preprint]{neurips_2024}

\usepackage{graphicx} 
\usepackage[utf8]{inputenc} % allow utf-8 input
\usepackage[T1]{fontenc}    % use 8-bit T1 fonts
\usepackage{hyperref}       % hyperlinks
\usepackage{url}            % simple URL typesetting
\usepackage{booktabs}       % professional-quality tables
\usepackage{amsfonts}       % blackboard math symbols
\usepackage{nicefrac}       % compact symbols for 1/2, etc.
\usepackage{microtype}      % microtypography
\usepackage{xcolor}         % colors
\usepackage{subcaption}
\usepackage{float} % Add this package for the [!htbp] option.

\title{PreND: Enhancing Intrinsic Motivation in Reinforcement Learning through Pre-trained Network Distillation}

\author{%
 \textbf{Mohammadamin Davoodabadi \quad\quad Negin Hashemi Dijujin \quad\quad Mahdieh Soleymani Baghshah }\\
 \\
 Department of Computer Engineering \\
 Sharif University of Technology \\
  \texttt{\{mohammadamin.davoodabadi, n.hashemi94, soleymani\}@sharif.edu}  \\
}

\begin{document}

\maketitle

\begin{abstract}
Intrinsic motivation, inspired by the psychology of developmental learning in infants, stimulates exploration in agents without relying solely on sparse external rewards. Existing methods in reinforcement learning like Random Network Distillation (RND) face significant limitations, including (1) relying on raw visual inputs, leading to a lack of meaningful representations, (2) the inability to build a robust latent space, (3) poor target network initialization and (4) rapid degradation of intrinsic rewards. In this paper, we introduce \textit{\textbf{Pre}-trained \textbf{N}etwork \textbf{D}istillation} (\textbf{PreND}), a novel approach to enhance intrinsic motivation in reinforcement learning (RL) by improving upon the widely used prediction-based method, RND. PreND addresses these challenges by incorporating pre-trained representation models into both the target and predictor networks, resulting in more meaningful and stable intrinsic rewards, while enhancing the representation learned by the model. We also tried simple but effective variants of the predictor network optimization by controlling the learning rate.
Through experiments on the Atari domain, we demonstrate that PreND significantly outperforms RND, offering a more robust intrinsic motivation signal that leads to better exploration, improving overall performance and sample efficiency. This research highlights the importance of target and predictor networks representation in prediction-based intrinsic motivation, setting a new direction for improving RL agents' learning efficiency in sparse reward environments.
\end{abstract}

\section{Introduction}

Reinforcement Learning (RL) has been acquired to solve many complex problems such as robot navigation \cite{miki2022learning},  stratospheric balloons navigation \cite{bellemare2020autonomous}, playing Go \cite{silver2016mastering}, and instruction-tuning large language models \cite{ouyang2022training,casper2023open}; however, by definition, it relies on reward signals from the environment which could be very cumbersome and labor-intensive to specify properly in the real-world tasks \cite{dulac2021challenges}. Many tasks especially in goal-oriented settings \cite{colas2022autotelicagentsintrinsicallymotivated} involve sparse rewards which provide weak learning signals for the agent and result in an ambiguous understanding of the environment. 

One of the methods proposed in the literature to overcome this challenge is \textit{intrinsic motivation} which is mainly inspired by the psychology and developmental learning of skills in babies \cite{aubret2019survey,doyleintrinsically}. Intrinsic motivation approaches involve different methods to improve exploration, learn diverse skills and enhance overall performance in complex environments using internally generated signals besides the external motivations \cite{aubret2019survey}. To this end, various approaches have been proposed which could be categorized into two main groups: 1) \textit{knowledge acquisition} and 2) \textit{skill learning}. Knowledge acquisition methods involve exploration-oriented methods based on prediction error \cite{burda2018exploration}, state novelty such as count-based methods \cite{tang2017exploration, pathak2019self}, and information gain \cite{houthooft2016vime}, or other methods based on empowerment \cite{choi2021variational} and state representation learning \cite{nachum2018near}. On the other hand, the skill learning category includes skill abstraction \cite{eysenbach2018diversity,sharma2019dynamics} and curriculum learning methods \cite{riedmiller2018learning}. Among all these categories, prediction-based methods and count-based methods have been mostly adopted throughout the literature due to their simplicity and promising performance \cite{hwang2023neuro}. However, count-based methods are not scalable in continuous and large state spaces while prediction-based methods handle these scenarios more efficiently \cite{aubret2019survey}.

One of the most important methods in prediction-based category is \textit{Random Network Distillation (RND)} \cite{burda2018exploration} which counts as the basis of many other studies \cite{yin2024random,guo2022byol,hwang2023neuro,badia2020never}. This method derives an intrinsic reward based on the prediction error between a random fixed function of states/observations and a learnable predictor network with the same input. The predictor network tries to learn the output of the fixed target network using a mean squared error objective; hence rewards the agent for visiting unfamiliar states that has seen less during training. Although this approach reduces uncertainty about the environment and seems suitable for exploring large state spaces, it is sensitive to target network initialization, produces rewards with low variance, and suffers from catastrophic forgetting throughout training \cite{pechavc2024self,hwang2023neuro}. Recent studies \cite{pechavc2024self,hwang2023neuro} have emphasized the shortcomings of RND. Specifically in \cite{pechavc2024self}, they propose self-supervised regularization of the target network to enhance representations of the target network. Target network representation also plays a crucial role as long-term memory keys in overcoming catastrophic forgetting, and also in specializing ensembles of curiosity modules over observation subspaces \cite{hwang2023neuro}.

In this paper, we emphasize the importance of target and predictor network representations in producing meaningful intrinsic rewards and improving the overall performance of the agent. Our analyses reveal the shortcomings of basic RND in generating meaningful and varied intrinsic rewards. We suggest that by changing design choices in random network distillation, network target initialization problems and low reward variance could be resolved. We explore using pre-trained models on the Atari domain from the recent Atari-PB benchmark \cite{kim2024investigating} as part of the target network initialization, and separate training speed for the overall policy and the predictor network to prevent its early overfitting to target network and low reward variance. Our experiments in Atari environment suggest effectiveness of these techniques. 

\begin{figure}[!htbp]
 \centering
 \includegraphics[width=0.53\textwidth]{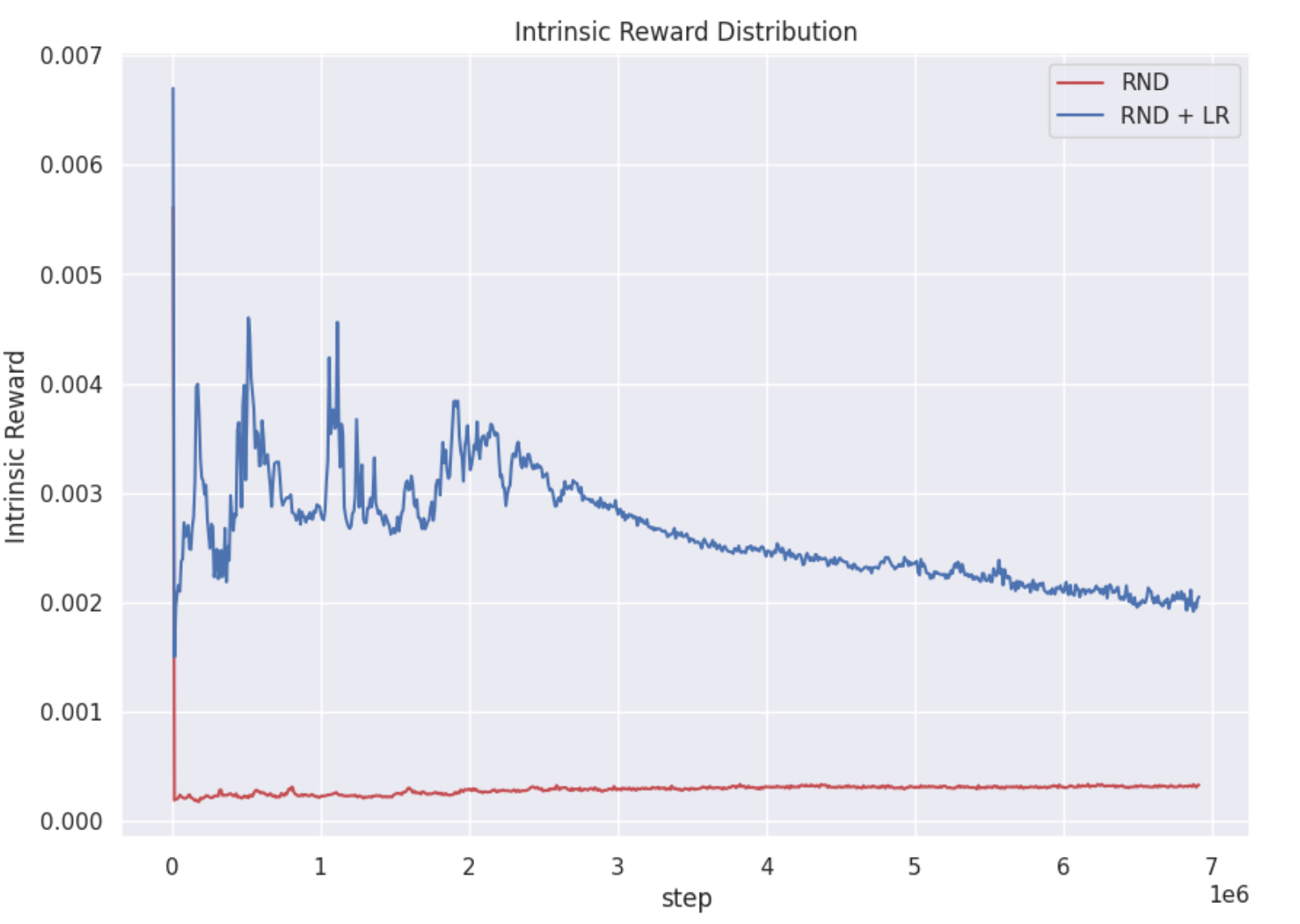}
 \caption{Effect of predictor's learning rate on intrinsic trend and oscillation of intrinsic reward} 
 \label{fig:int}
\end{figure}

\section{Method}
In this section, we present the methodology behind our proposed approach, \textit{\textbf{Pre}-trained \textbf{N}etwork \textbf{D}istillation} (\textbf{PreND}). As discussed, RND relies on the difference between the outputs of a target network and a predictor network to generate intrinsic motivation. However, it has the following problems:

\begin{enumerate}
    \item A random target network might not represent the semantics of the input observations/states in producing the intrinsic reward. This might happen due to poor latent space modeling in a random network. Similar observations might scatter through the latent space while different observations could be mapped to closer latent points. This can confuse the predictor and produce false surprise signals. We calculated the correlation between intrinsic rewards difference and state embedding distance for RND. The resulting number ($\approx 0.39$ on the scale of $[-1, 1]$) indicates a weak correlation \cite{evans1996straightforward}, leaving room for improvement (See Appendix Figure \ref{fig:hm} for visualizations).
    \item The intrinsic reward drops quickly after very few initial iterations, leading to poor reward variance (Figure \ref{fig:int}). Similar rewards make it more difficult for the agent to discriminate between states while the whole point of prediction-based intrinsic reward is to help the agent visit under-explored states by giving it relatively considerable rewards. Previous research has noted that the lack of variance in the predictor’s output leads to a diminished motivational signal \cite{pechavc2024self}, primarily due to the predictor network's very fast adaptation to the frozen target model.
\end{enumerate}

To this end, we propose PreND, consisting of the following techniques to improve the abovementioned weaknesses:

\textbf{Pre-trained Feature Extractor:} Using pre-trained models for the target network, we can improve the fixed random target network problem. These models can provide a meaningful feature space that preserves the semantics of observations and can induce relatively suitable prediction errors among observations/states. This approach could also alleviate the robustness of networks in confronting noisy observations with distractor objects through meaningful learned representations. 
    
    More specifically, in our experiments on the Atari environment, we propose to use the domain-specific pre-trained backbone model from the Atari-PB benchmark to extract high-level representations. We then employed a randomly initialized network, similar to the neck model architecture from Atari-PB (a transformer-based random fixed network), as the target network, while a similar learnable model was selected for the predictor. This ensures that the predictor and target are complex enough to understand meaningful features of the input, and the predictor is capable of modeling the output of the target network (See Appendix \ref{modelarch} for more information).

     Our target network maintains the crucial property of keeping distant frames far apart in latent space while bringing consecutive frames closer. Unlike RND, where the predictor also had to learn the spatial and temporal features, PreND ensures that the predictor focuses solely on predicting the reward, resulting in a more robust motivational signal.
    
\textbf{Slower Predictor Optimization:} To alleviate the fast degradation of intrinsic reward, we suggest lowering the optimization speed in the predictor network. Normally, RND uses the same optimizer for both predictor and value/policy heads. Since fitting RL components are known to be more sample-inefficient compared to supervised tasks, an equal optimization speed increases the chance of collapse in intrinsic reward. By changing the pace via learning rate, similar to previous studies \cite{madan2021fast}, we hope to improve intrinsic reward patterns during training.

To summarize, our experiments demonstrate that PreND outperforms RND by providing a richer representation of the input throughout the training, leading to better correlation between changes in input and changes in reward. This alignment is critical for prediction-based intrinsic motivation methods, and PreND offers an effective solution for addressing the limitations of RND.

\section{Experiments}
We evaluated our approach on two Atari environments: \texttt{Boxing}, and \texttt{Riverraid} (See Appendix Figure \ref{fig:games}). These environments were chosen because they share several characteristics that are particularly relevant to our study. Each game contains highly irrelevant features, such as detailed backgrounds, which our pre-trained representation model should ideally ignore. Moreover, the objects in these games are relatively large, which provides a more favorable setting for RND to compete fairly. In contrast, RND's performance typically suffers in environments with small moving objects, like the ball in Pong \cite{hwang2023neuro}, as its random target network struggles to capture the subtle movements.

\begin{figure}[!htbp]
 \begin{subfigure}[b]{0.48\textwidth}
 \centering
 \includegraphics[width=\textwidth]{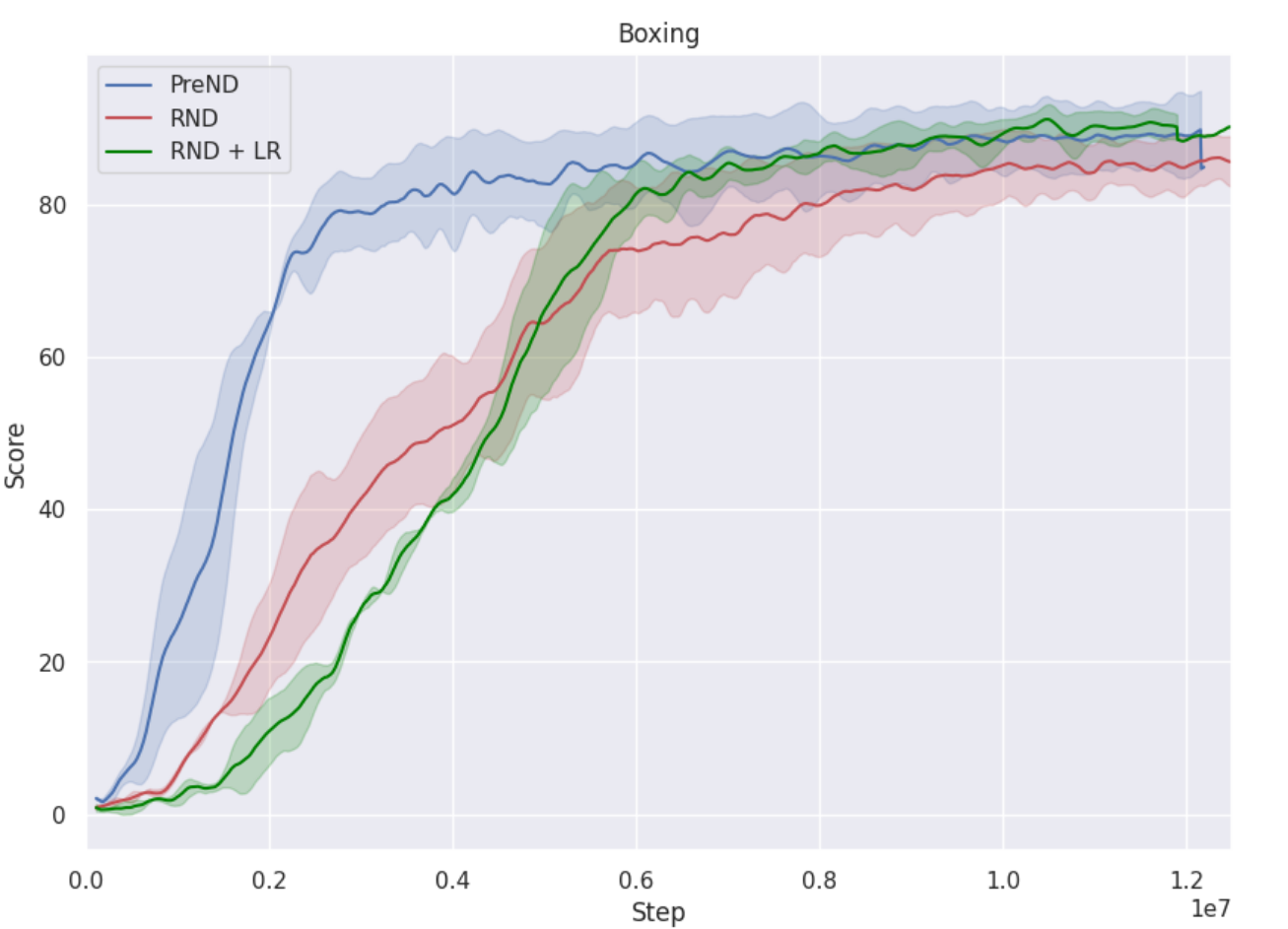}
 % \caption{a}
 \label{fig:res1}
 \end{subfigure}
 \hfill
 \begin{subfigure}[b]{0.50\textwidth}
 \centering
 \includegraphics[width=\textwidth]{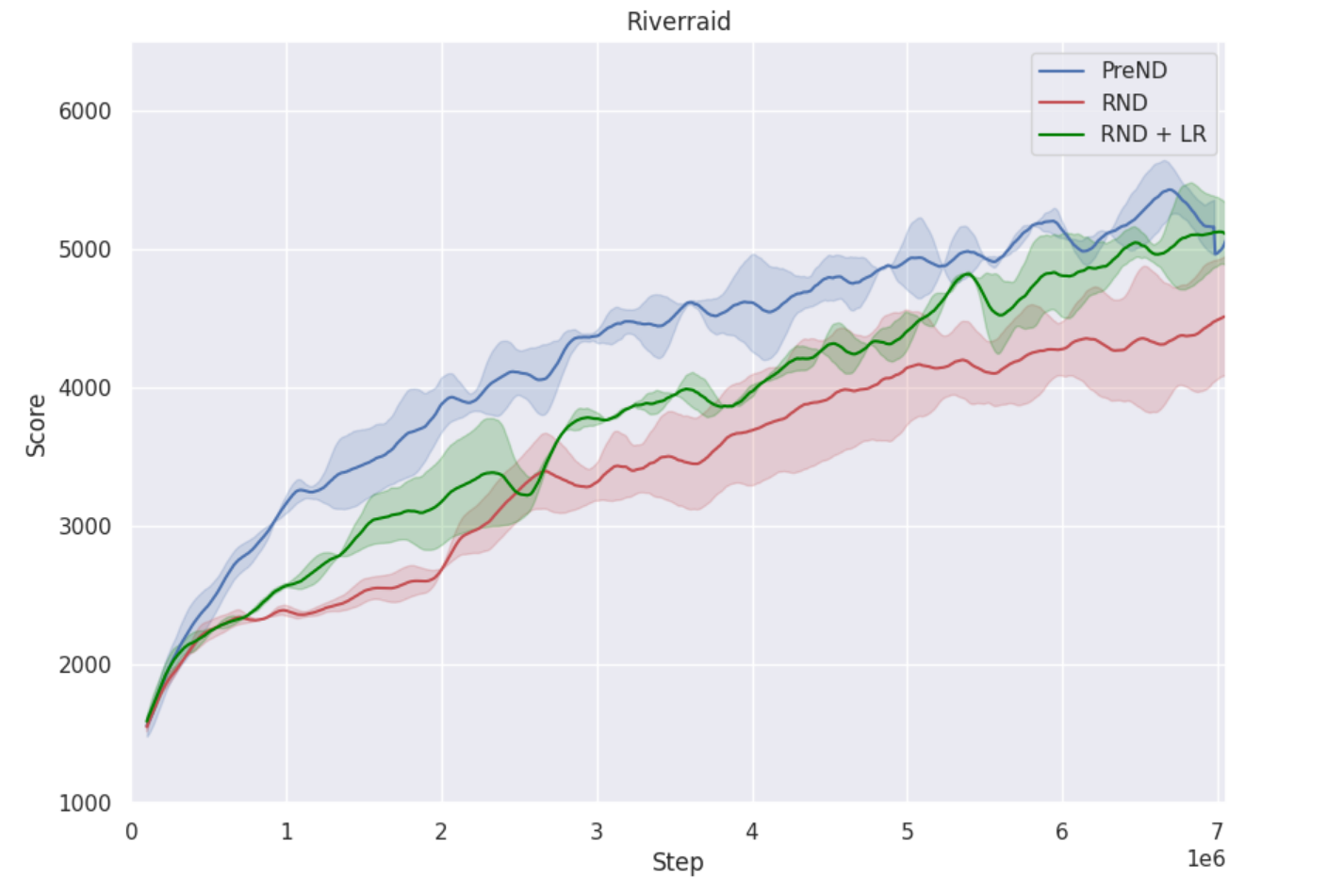}
 % \caption{b}
 \label{fig:res2}
 \end{subfigure}
 
 \caption{Game score comparison on \texttt{Boxing} and \texttt{Riverraid}. PreND works better in the low data regimes, showing its sample efficiency. It is also the best model among the settings at the end of the training steps.} 
 \label{fig:res}
\end{figure}

For each environment, we ran our experiments with two different seeds and eight parallel episodes, and compared the following three models:
\begin{itemize}
    \item \textbf{RND}\cite{burda2018exploration}: This was our main competitor, offering a prediction-based intrinsic reward to drive exploration. RND utilizes CNNs as the predictor and target networks.
    \item \textbf{RND + LR}: A variant of RND where we adjusted the learning rate to prevent the predictor network from overcoming the target network. We consider a separate optimization process for fitting the predictor network and multiply its learning rate by $0.01$.
    \item \textbf{PreND}: Our proposed method, leverages a pretrained representation model for intrinsic reward generation. We use the pre-trained ResNet-50 backbone and SiamMAE neck from Arati-PB \cite{kim2024investigating} as the target and predictor network structure. Backbone (which is pre-trained on several Atari games) retains its pre-trained weights and acts as a feature encoder for observations. We randomly initialize the neck for both the target and predictor networks to prevent game-specific bias of the pre-trained weights. The intrinsic reward in calculated over the neck's output with a size of 512, as in RND.
\end{itemize}
During our experimentation, we also explored several modifications and variants aimed at further improving RND. These included reducing the capacity of the predictor network by removing additional layers and incorporating techniques such as spectral normalization (commonly used in GAN variants \cite{miyato2018spectralnormalizationgenerativeadversarial}) to the predictor hoping to make its learning process slower and improve the intrinsic reward variance. However, none of these modifications led to improvements. The main results have been shown in Figure \ref{fig:res}.

\section{Conclusion}
In this paper, we introduced Pre-trained Network Distillation, or PreND for short, a novel approach to improving intrinsic motivation in reinforcement learning by leveraging pre-trained representation models. Through experiments in Atari games, we demonstrated that PreND performs better than both Random Network Distillation (RND) and a variant of RND with a modified learning rate. While the modified RND showed improvements over the baseline by addressing some of the RND’s inherent issues, PreND’s use of richer input representations and pre-trained models seems even more effective. The stable and informative reward signal generated by PreND allowed for better exploration and learning efficiency, confirming that target and predictor network representations play a critical role in producing meaningful intrinsic rewards.

Our results suggest the probable benefits of PreND in the Atari domain, but future work could extend it to more complex environments like DMLab or robotics and explore its use with model-based RL algorithms in more extensive trials. Additionally, experimenting with lighter pre-trained models, such as ResNet-18 instead of ResNet-50, could offer more computational efficiency while maintaining effectiveness.

% \bibliographystyle{plain}
% \bibliography{neurips_2024} 

\appendix

\section*{Appendix}

\begin{figure}[!htbp]
 \begin{subfigure}[b]{0.49\textwidth}
 \centering
 \includegraphics[width=\textwidth]{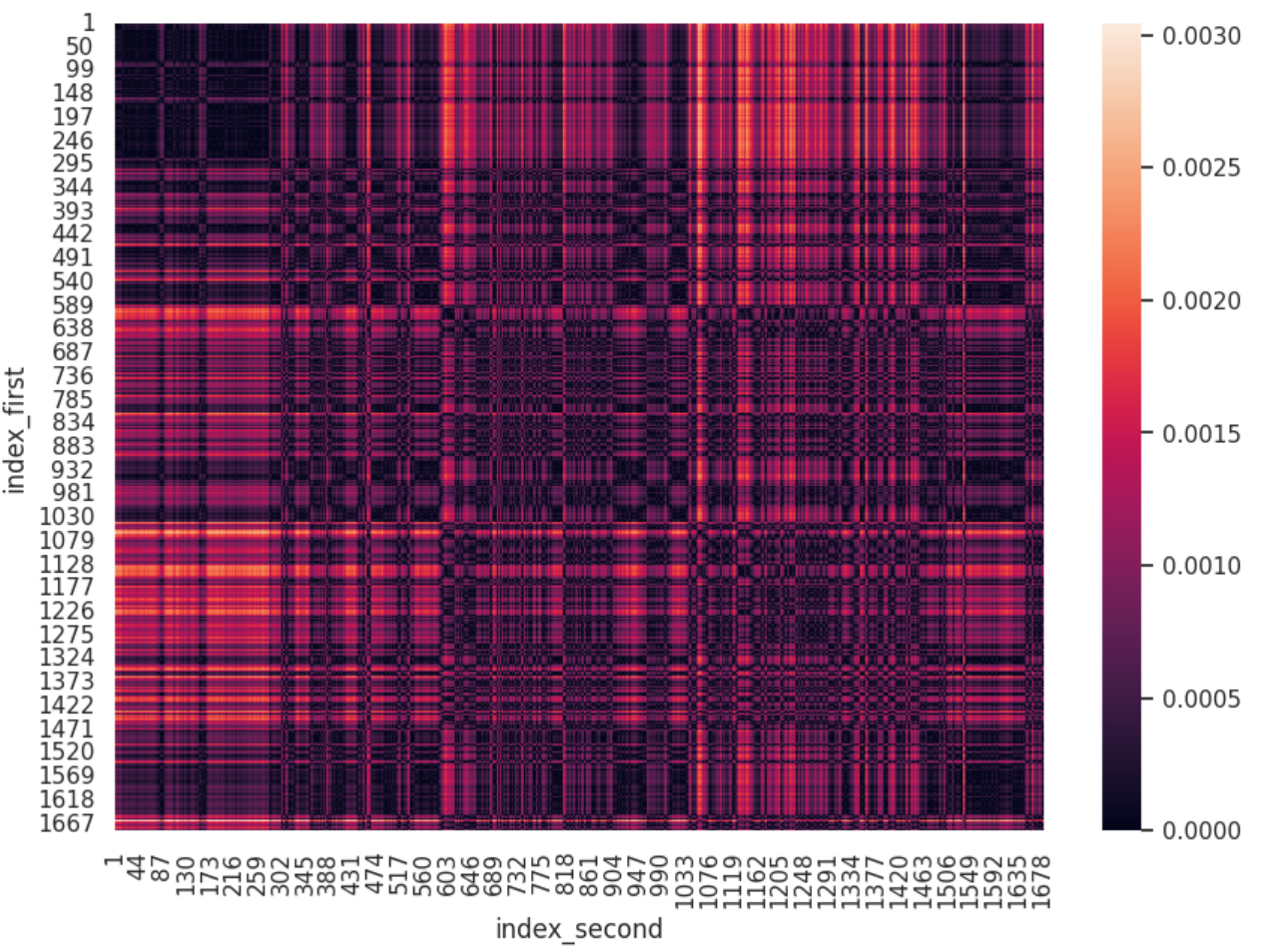}
 \caption{Pairwise intrinsic reward difference heatmap}
 \label{fig:hmrew}
 \end{subfigure}
 \hfill
 \begin{subfigure}[b]{0.47\textwidth}
 \centering
 \includegraphics[width=\textwidth]{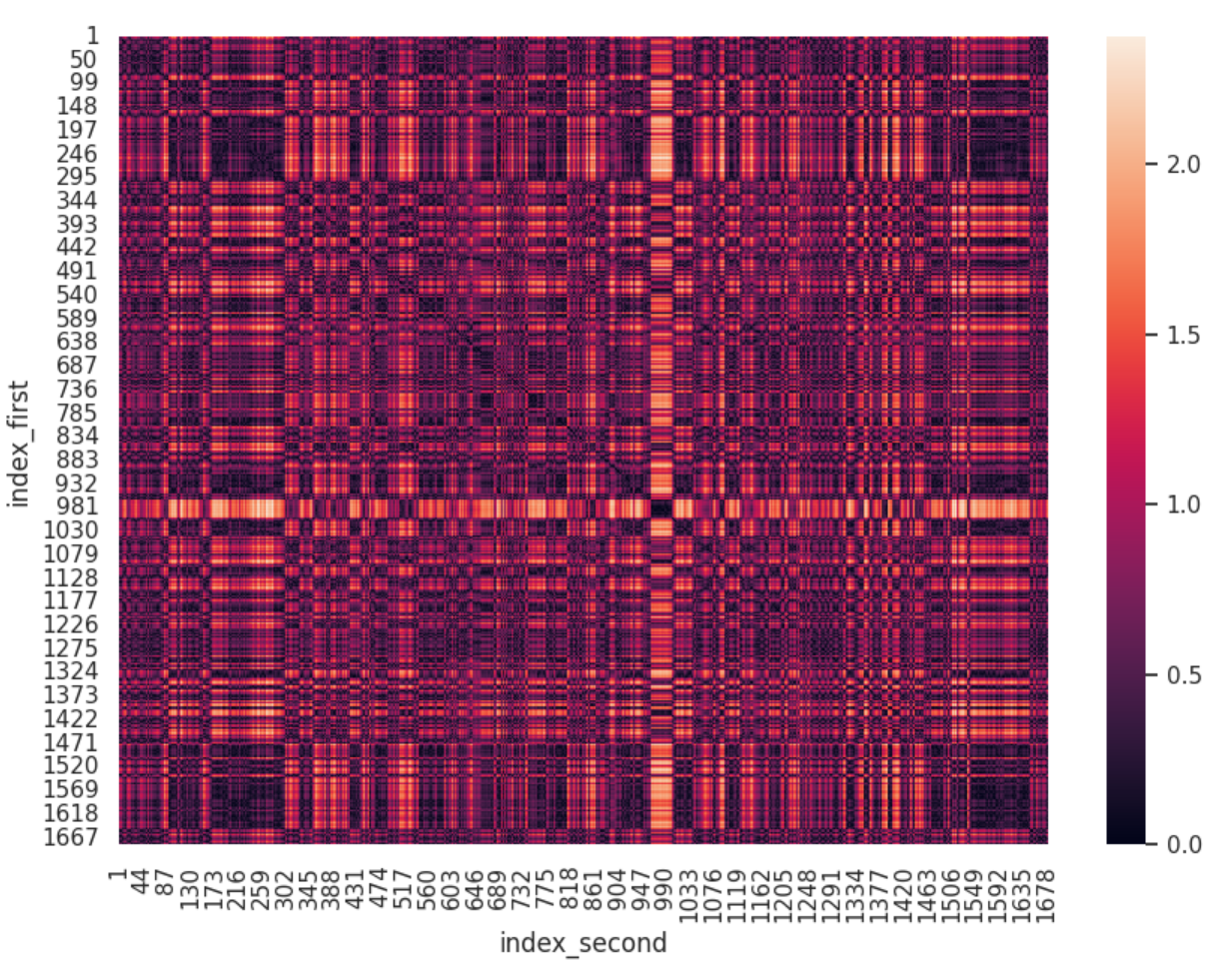}
 \caption{Pairwise observation distance heatmap}
 \label{fig:hmobs}
 \end{subfigure}
 \caption{Heatmaps of pairwise training variables; The correlation between these two is $\approx 0.39$ on the scale of $[-1, 1]$. X and Y axes are both the number of training iterations. In both a) and b), the brighter cells indicate higher values (More distance). While it's essential for prediction-based methods to capture the change of states in the form of an intrinsic reward, we can see that there is no strong relation between what RND sees as input and what it produces as an intrinsic reward. } 
  \label{fig:hm}
\end{figure}

\section{Related Work}
Intrinsic motivation aims to improve exploration at every timestep. Exploration in learning agents helps them acquire various skills required to accomplish diverse tasks \cite{chentanez2004intrinsically}. Intrinsic motivation in reinforcement learning is inspired by the psychology and developmental learning of skills in babies and helps tackle several RL challenges \cite{aubret2019survey} such as sparse rewards \cite{burda2018exploration}, representation learning \cite{binas2019journey}, skill learning \cite{eysenbach2018diversity,sharma2019dynamics}, and curriculum learning \cite{florensa2017reverse}. There are different approaches to improving exploration using intrinsic motivation which mainly include count-based \cite{tang2017exploration,ecoffet2021first} and prediction-based auxiliary rewards \cite{burda2018exploration,hwang2023neuro,guo2022byol,badia2020never}.
Count-based approaches are unsuitable for large or continuous state spaces. Some studies try using pseudo-counts and neural density modules to alleviate the scalability problem in count-based methods \cite{ostrovski2017count,machado2020count,martin2017count}. On the other hand, prediction-based approaches which are mainly represented by RND \cite{burda2018exploration} could suffer from inappropriate target network initialization, high surprise variance, and early degradation of surprise overtime \cite{pechavc2024self}. In \cite{pechavc2024self}, the authors propose improving target network weights using self-supervised regularization techniques. Another approach might rely on using pre-trained representations in the target network. Such representations are abundantly available for various real-world tasks employing large pre-trained vision models \cite{radford2021learning,dosovitskiy2020image}. Also in more narrow environments such as Atari, domain-specific pre-trained models \cite{kim2024investigating} offer a chance to examine this hypothesis in the context of prediction-based models such as RND. In this study, we incorporate ResNet-based backbone networks in prediction and target networks to improve the quality of surprise.

\section{Experiments Details}

We have used the codebase from \cite{hwang2023neuro} to run RND and implemented our method on top of it using the codebase of \cite{kim2024investigating}. We ran our reported experiments on two Nvidia GeForce RTX 3090 GPUs and each run took nearly 24 hours.

\section{Model Architecture}
\label{modelarch}

\textbf{Backbone:} The backbone network is a spatial feature extractor, independent of the game, based on ResNet-50 with group normalization. It processes a (4, 84, 84) input and outputs a (2048, 6, 6) feature map, encoding essential features as the input of the target and predictor network of PreND. 

\textbf{Neck:} The neck transforms the feature map from the backbone into a 512-dimensional vector by applying spatial pooling, instance normalization, and a 2-layer MLP. It includes game-specific spatial embedding. In the SiamMAE, the output from the neck is further processed by the transformer decoder, which uses cross-attention between the current and masked future frames to reconstruct the image. As \cite{razzhigaev2024transformersecretlylinear} said, the transformers maintain the linear property; which is important to ensure that the model keeps the similar frames close to each other in the latent space and the dissimilar ones far from each other.

\begin{figure}[!htbp]
 \begin{subfigure}[b]{0.49\textwidth}
 \centering
 \includegraphics[width=\textwidth]{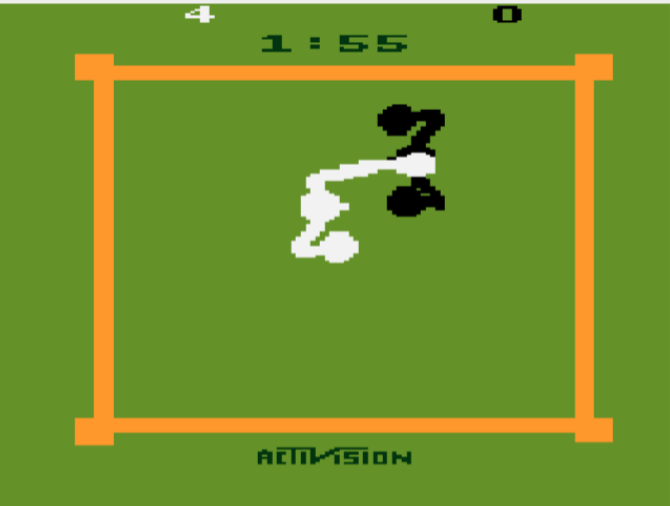}
 \caption{Boxing}
 \label{fig:boxinggame}
 \end{subfigure}
 \hfill
 \begin{subfigure}[b]{0.47\textwidth}
 \centering
 \includegraphics[width=\textwidth]{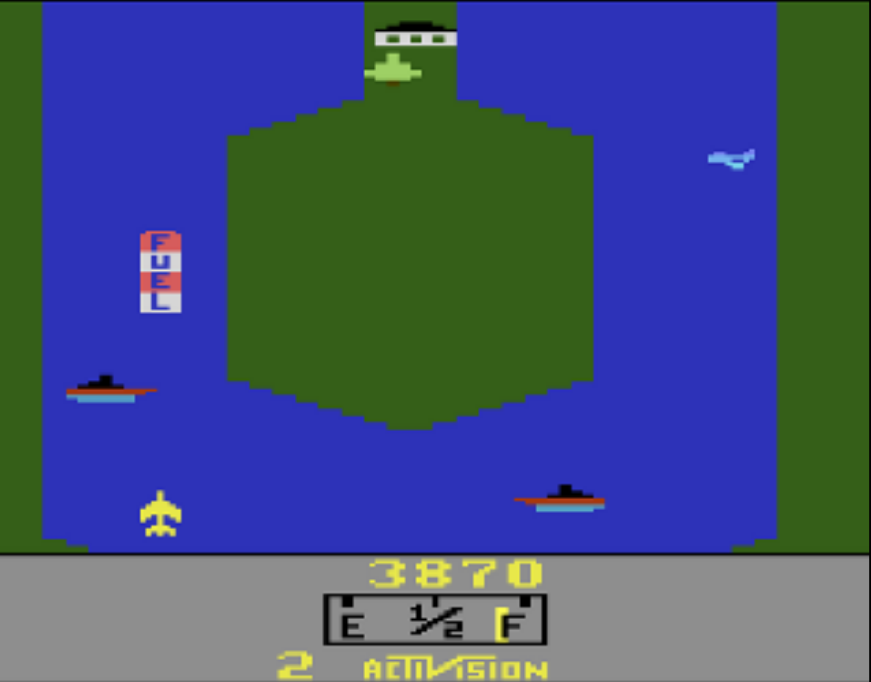}
 \caption{River Raid}
 \label{fig:rivergame}
 \end{subfigure}
 \caption{Atari games} 
 
 \label{fig:games}
\end{figure}

\end{document}